\documentclass[11pt]{article}

\usepackage[preprint]{acl}

\usepackage{times}
\usepackage{latexsym}

\usepackage[T1]{fontenc}

\usepackage[utf8]{inputenc}

\usepackage{microtype}

\usepackage{inconsolata}

\usepackage{graphicx}

\usepackage{enumitem}
\usepackage{amsmath}
\usepackage{amssymb}

\usepackage{enumitem}
\usepackage{graphicx}
\usepackage{amssymb}
\usepackage{amsmath}
\usepackage{multicol}
\usepackage{multirow}
\usepackage{array}
\usepackage{algorithmic}
\usepackage{algorithm}
\usepackage{subcaption}
\usepackage{wrapfig}
\usepackage{microtype}
\usepackage{graphicx}
\usepackage{subcaption}
\usepackage{booktabs} 
\usepackage{amssymb,mathrsfs}
\usepackage{enumitem}
\usepackage{wrapfig}
\usepackage{subcaption}
\usepackage{tabularx}
\usepackage{placeins}
\usepackage{xcolor}         
\usepackage{array}
\usepackage{multicol}
\usepackage{multirow}
\newcommand{\ours}[0]{{{E$^2$GraphRAG}}}

\title{E$^2$GraphRAG: Advancing the Empirical Pareto Frontier\\ in Efficiency and Effectiveness for Graph-based RAG}

\author{  Yibo Zhao$^{1}$ \quad
  Jiapeng Zhu$^{1}$ \quad
  Jianxiang Yu$^{1}$\\
  {\bf Ye Guo}$^{2}$\quad
  {\bf Kangkang He}$^{2}$\quad
  {\bf Xiang Li}$^{1}$\thanks{Corresponding Author: \texttt{xiangli@dase.ecnu.edu.cn}} \\
  $^1$School of Data Science and Engineering, East China Normal University \\
  $^2$China Baowu Group}

\begin{document}
\maketitle
\begin{abstract}
Graph-based RAG methods such as GraphRAG achieve strong global understanding by constructing hierarchical entity graphs, yet their LLM-heavy indexing and pre-defined query workflows hinder efficient deployment. We propose \ours, a streamlined graph-based RAG framework that advances the empirical Pareto frontier of \underline{E}fficiency and  \underline{E}ffectiveness. In the indexing stage, \ours \ utilizes large language models to generate a hierarchical summary tree and lightweight NLP tools to construct an entity graph from document chunks, with bidirectional indexes linking entities and chunks for efficient lookup. In the retrieval stage, the graph structure filters related entities, while the bidirectional indexes map these entities to their corresponding chunks, supporting an adaptive mechanism that dynamically switches between local and global modes. Experiments on three long-document QA datasets with two backbone models show that \ours \ achieves up to $10\times$ faster indexing than GraphRAG while maintaining comparable QA performance, improving the observed indexing efficiency--QA
effectiveness frontier under the evaluated settings. Our code is publicly available at \url{https://github.com/YiboZhao624/E-2GraphRAG}.
\end{abstract}

\section{Introduction}

With the continuous advancement, large language models (LLMs)~\citep{flashattention, KVcache, attention} have become a cornerstone in NLP, which have been widely applied in tasks such as text summarization~\citep{summary-1,summary-2}, machine translation~\cite{translation-1, translation-2}, and question answering~\citep{QA-3, QA-1, QA-2}. However, they still face limitations, including hallucinations~\citep{hallucination-2, hallucination-1, hallucination-survey, hallucination-solve} and a lack of domain-specific knowledge~\citep{domain-knowledge-3, domain-knowledge-2, domain-knowledge-1, domain-knowledge-survey}. 
To address these issues, Retrieval-Augmented Generation (RAG) has been proposed~\citep{RAG-survey-1, RAG-survey-2, RAG-0}. 
By retrieving relevant knowledge from external sources and leveraging the in-context learning capabilities, RAG allows models to integrate timely and domain-specific information, mitigating knowledge gaps.

Traditional RAG methods typically retrieve only a small set of chunks from original documents as supplemental knowledge. However, this limited context could be insufficient for providing the model with a comprehensive and global understanding of the knowledge base, such as understanding and summarizing a character's personality transformation, as in NovelQA~\citep{NovelQA}. Consider the novel \textit{Harry Potter and the Prisoner of Azkaban} and the question: ``\textit{Peter Pettigrew used to be positive and finally becomes a negative one. Tell in one sentence what marks this character's change}.'' Traditional RAG methods typically retrieve only a few isolated chunks about Peter Pettigrew, whereas answering this question requires a holistic understanding of his entire character arc.

To address the fragmented evidence issue, existing
state-of-the-art methods, including RAPTOR~\citep{raptor}, GraphRAG~\citep{graphrag}, and LightRAG~\citep{lightrag}, adopt an \emph{{LLM-assisted structured} indexing-and-retrieval paradigm}: they first use LLMs to index the documents into tree- or graph-based structures\footnote{Since tree is a special form of graph, we uniformly use graph-based RAG in this paper.} and then retrieve on these structured data.
While hierarchical trees, constructed by recursively merging text chunks, provide global understanding, they are limited in capturing fine-grained knowledge, such as entities and their relations. Entity graphs, on the other hand, enable the extraction and integration of such fine-grained knowledge across chunks, but they heavily rely on LLM-based entity and relation extraction, which is computationally expensive.

To summarize, existing approaches face three major challenges.
First, efficiency remains the primary bottleneck: although several efforts~\citep{fast-graphrag, lightrag} have attempted to reduce computational overhead, indexing and retrieval are still far from optimal.
Second, most methods rely exclusively on either tree or graph structures to organize raw, lengthy text. While each structure has its advantages, their joint integration has not been thoroughly investigated.
Third, in the retrieval stage, some approaches~\citep{graphrag, lightrag} depend on manually pre-defined query modes (e.g., local or global), resulting in limited flexibility.
Therefore,
a research question naturally emerges:
{\textit{
Is it possible to design a graph-based RAG model that advances the efficiency-effectiveness Pareto frontier while adaptively answering queries at varying granularities?
}}

In this paper, we streamline graph-based RAG for high efficiency and effectiveness, and propose the \ours\ framework, which combines the strengths of both tree and graph structures.
Specifically, we first recursively merge and summarize text chunks to construct a \textbf{hierarchical summary tree}, enabling multi-granularity summarization of raw text.
To integrate fine-grained knowledge from dispersed chunks, we also construct a concise \textbf{entity graph}. 
Rather than relying on LLMs for entity extraction, we employ the lightweight NLP tools such as SpaCy~\citep{spacy}, {HanLP}~\cite{hanlp}, and define relations based on entity co-occurrence within a sentence.
We further construct bidirectional entity-to-chunk and chunk-to-entity indexes to bridge the entity graph and the summary tree, facilitating efficient lookup during subsequent retrieval.
In the retrieval stage, we introduce an efficient adaptive strategy that leverages the entity graph to select between local and global query modes: queries whose entities are densely connected are processed locally, while others fall back to global retrieval.
This mechanism models structural relationships among entities explicitly, eliminating the need for manually predefined query modes and enabling more efficient and flexible retrieval for diverse query types.

In summary, our contributions are threefold:

\begin{itemize}[leftmargin=*, itemsep=0pt, topsep=0.2pt]
\item We propose \ours, a novel framework that integrates a summary tree and an entity graph via bidirectional indexes, bringing new insights into lightweight graph-based RAG indexing by constructing the entity graph without LLMs.
\item We design graph-driven adaptive retrieval that automatically selects between local and global modes, eliminating manual query presets.
\item We conduct extensive experiments showing that \ours\ achieves up to $10\times$ faster indexing than GraphRAG while maintaining competitive QA performance, advancing the Pareto frontier.
\end{itemize}

\section{Related Work}


RAG has been extensively studied, where most existing methods fall into two main categories based on the type of external knowledge source. 
Most approaches~\citep{selfrag} rely on unstructured textual knowledge bases, which are easy to organize and adaptable to various tasks, but often lack a global and structured understanding of the content. 
Others utilize {structured knowledge graphs}~\citep{gretriever, dalk, thinkongraph}, which support multi-hop reasoning and information aggregation.
However, building high-quality knowledge graphs typically requires substantial expert efforts and is difficult to scale.

GraphRAG~\citep{graphrag} is the first method to automatically construct knowledge graphs from raw text, supporting both local and global queries, which attracts considerable attention~\citep{graphrag-survey-1, graphrag-survery-2}. 
While it achieves strong effectiveness through community-based summarization, its indexing and retrieval incur substantial costs due to numerous LLM calls and verbose structured outputs. 
Recent advancements can be categorized into two distinct lines based on the semantic richness of the constructed graphs.
The first line focuses on extracting graphs enriched with semantic information, where nodes and edges possess explicit textual descriptions.
LightRAG~\citep{lightrag} and FastGraphRAG~\citep{fast-graphrag} eliminate community summarization: LightRAG directly extracts low- and high-level nodes from each chunk, while FastGraphRAG leverages PageRank for query-time aggregation. 
However, these methods still rely on LLMs to produce structured outputs. 
In contrast, the second line utilizes graphs primarily as lightweight retrieval scaffolds composed of entities and relations without detailed descriptions. HippoRAG~\citep{hipporag} and HippoRAG2~\citep{hipporag2} apply PageRank-style propagation to reweight query-relevant entities, improving retrieval with limited indexing overhead, but they do not perform community-level summarization. Critically, these methods leverage graph topology merely to enhance embedding similarity for accurate retrieval, acting as an auxiliary ranking signal rather than performing fully global graph reasoning.
In concurrent works, Youtu-GraphRAG~\cite{youtugraphrag} and LinearRAG~\citep{linearrag} further optimize the token consumption and extraction efficiency.
RAPTOR~\cite{raptor} adopts a hierarchical summary tree to efficiently preserve multi-level context, while trading off detailed entity-level reasoning. 
Overall, existing methods illustrate a tension between efficiency and effectiveness: rich semantic graphs prioritize global reasoning but incur high computational costs, whereas simplified retrieval scaffolds improve efficiency at the expense of deep semantic comprehension. Yet, beyond these graph-centric trade-offs, methods that simultaneously leverage both graph and tree structures remain largely underexplored.

\begin{figure*}
\centering
\includegraphics[width=0.85\textwidth]{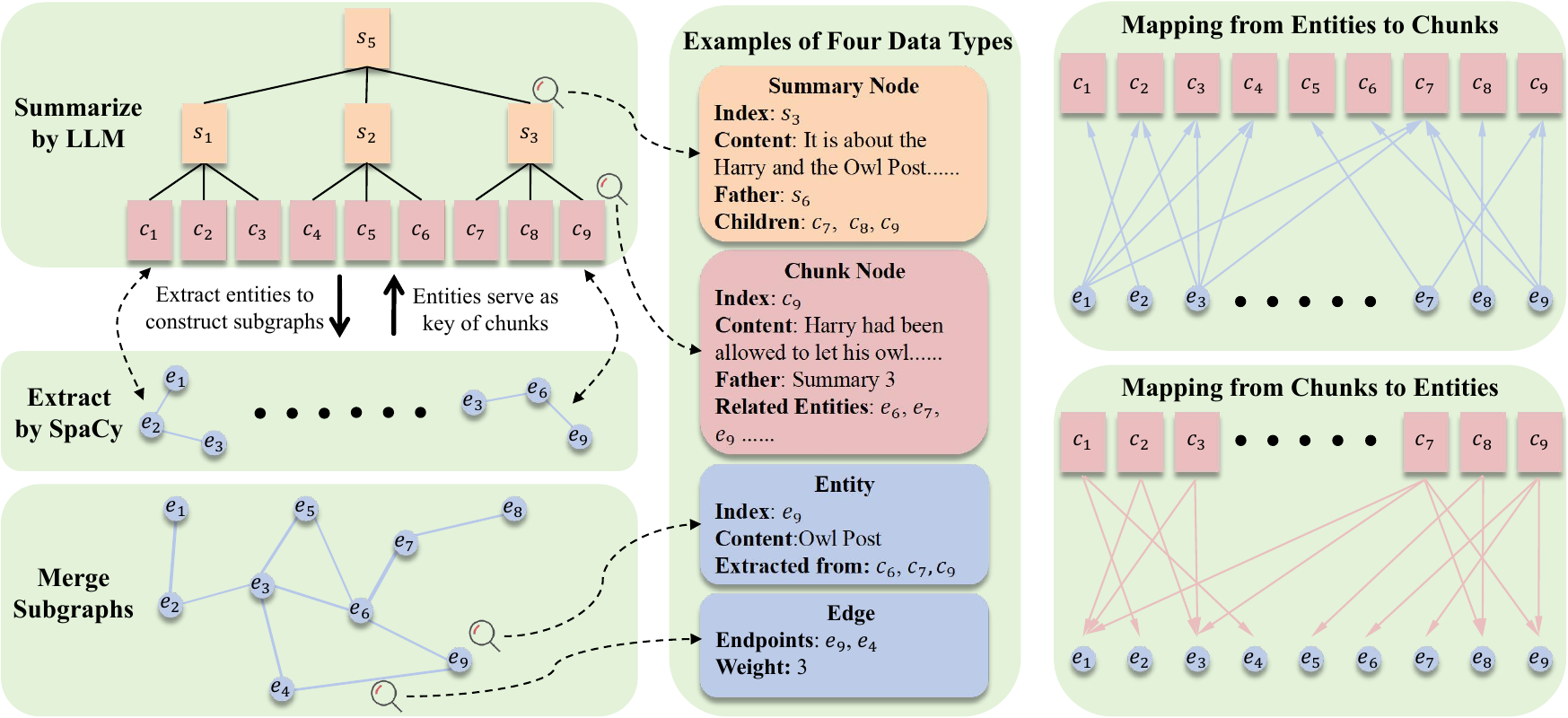}
\caption{Overview of the indexing stage of \ours. The left part shows the indexing tasks, the center presents the four data structures, and the right part displays the two constructed indexes.} \label{fig:index}
\end{figure*}

{Motivated by this tension between efficiency and effectiveness, \ours\ constructs an entity co-occurrence graph using lightweight NLP tools and builds a hierarchical summary tree to retain multi-granularity context.}

\section{Method}\label{sec:method}


Our approach consists of two main stages: \textbf{indexing} and \textbf{retrieval}. For our task, we first introduce some symbolic definitions to facilitate clearer explanations in subsequent sections. As input, we use $D$ to represent the document serving as the knowledge base, $q$ to denote the query, and $k$ to denote the maximum number of chunks retrieved.

\subsection{Indexing Stage}

As in standard RAG indexing, we first split each document into $n$ chunks. 
We tokenize the document using the tokenizer corresponding to the model used in the subsequent summarization task, and divide it into chunks of 1200 tokens each, with an overlap of 100 tokens between adjacent chunks to mitigate the semantic loss caused by potential sentence fragmentation. 
The resulting chunked document is denoted as $D = \{c_1, c_2,\cdots, c_n\}$.
Then, as illustrated in Figure~\ref{fig:index}, the indexing stage comprises two main tasks: construction of a \textbf{summary tree} and extraction of an \textbf{entity graph}. 
To enhance subsequent retrieval, we further introduce two types of indexes that establish many-to-many mappings between the tree and the graph.

For the summary tree construction,
we preserve the original chunk order and employ an LLM to summarize every consecutive group of $g$ chunks. 
Notably, since most modern LLMs have been extensively trained on text summarization tasks~\citep{LLMSummary2, LLMSummary1}, we adopt a minimal prompting strategy\textemdash providing only task instructions without lengthy few-shot examples, 
\textemdash thereby improving indexing efficiency.
Once all the original chunks have been summarized, the resulting summaries are treated as a new sequence of inputs.
This recursive summarization process continues, grouping every $g$ summaries at each level, until only $g$ or fewer segments remain.
Through the above procedure, the raw document is transformed into a tree structure, where the leaf nodes correspond to chunks, and the intermediate or root nodes correspond to the summaries.
Nodes closer to the root contain more global and abstract information, while those closer to the leaves retain more detailed and specific content.
We then utilize a pretrained embedding model to encode all chunks and summaries, storing the resulting vectors using \textit{Faiss}~\citep{faiss} to enable efficient dense retrieval.
Formally, we denote the summary tree as $T = \{c_1, \cdots,c_n, s_1,\cdots,s_o\}$, where each chunk $c_i$ and summary $s_i$ is a node in the tree. 

For the entity graph extraction task, instead of relying on LLMs to extract entities and relations as in GraphRAG-style approaches, 
we employ lightweight entity extractors, such as SpaCy~\citep{spacy}, NLTK~\citep{nltk}, and fine-tuned BERT~\citep{BERTNER}, which are significantly more efficient than LLMs for large-scale information extraction.
In particular, we extract named entities and common nouns (as nouns often indicate potential entities), and uniformly refer to them as \textit{entities} hereafter. Formally, for each chunk $c_i$, we denote the extracted entities as $\mathcal E_{c_i} = \{e_{c_i}^1, \cdots,e_{c_i}^m\}$, where $m$ is the number of entities identified in chunk $c_i$.
After extracting entities, we construct an undirected weighted edge between any two entities that co-occur within the same sentence, where the edge weight reflects their sentence-level co-occurrence frequency. This results in a subgraph $\mathcal G_{c_i}$ for each chunk $c_i$, which captures the relations among entities mentioned within the chunk and allows us to construct associations between entities and chunks. 
To support efficient retrieval, we build two one-to-many indexes to link entities and chunks, thereby capturing the many-to-many relations between them. The entity-to-chunk index, $I_{e\rightarrow c}(\cdot)$, maps each entity to the set of chunks where it appears. The chunk-to-entity index, $I_{c\rightarrow e}(\cdot)$, records the entities extracted from each chunk. These two indexes establish a many-to-many mapping between the entities in the entity graph and the chunks in the summary tree, facilitating the subsequent entity-aware retrieval stage.
For the entire document, we merge all chunk-level subgraphs into a single graph $\mathcal G$, where identical entities are unified and edges with the same source and target entities have their weights summed.
Since some entities appear in multiple chunks, this merging allows the graph to capture the co-occurrence relationships among entities across the document.


In conclusion, as illustrated in Figure~\ref{fig:index}, our method involves four types of data in two data structures: summary nodes and original chunk nodes in the tree, along with entities and weighted edges in the graph. 
In addition, our method relies on two key indexes, chunk-to-entity index $I_{c \rightarrow e}(\cdot)$ and entity-to-chunk index $I_{e \rightarrow c}(\cdot)$, which bridge the tree and the graph. These indexes enable efficient mapping from a chunk to its associated entities, and from an entity to the chunks in which it appears, respectively, thereby facilitating subsequent retrieval.

\begin{figure*}
\centering
{
\includegraphics[width=0.85\textwidth]{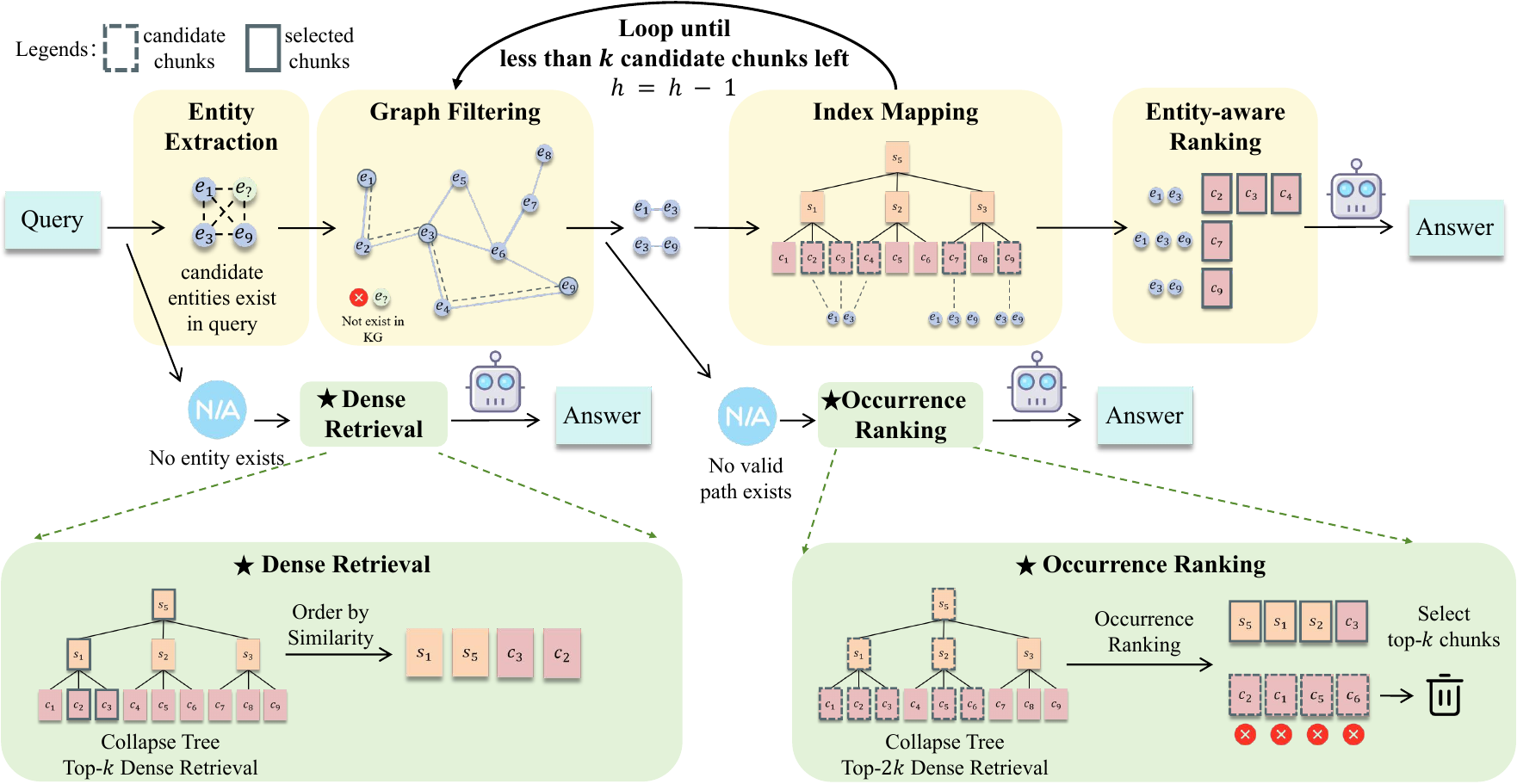}}
\caption{The retrieval stage of E$^2$GraphRAG. Operations belonging to the local retrieval are highlighted in light yellow, while those for global retrieval are highlighted in light green and marked with a $\bigstar$.} \label{fig:query}
\end{figure*}

\subsection{Retrieval Stage}

In the retrieval stage, previous work faces two main challenges: (1) global queries heavily rely on LLMs, resulting in high retrieval latency, and (2) the retrieval hierarchy and methods often require manual specification, introducing additional hyperparameters that are difficult to optimize. 
To address these issues, we first introduce a novel retrieval mechanism that adaptively selects between global and local retrieval when specific logical conditions are met. Then, we rank and format the retrieved chunks, thereby enhancing the LLM.
To clearly distinguish between the two adaptively selected retrieval modes, we highlight 
\textbf{global retrieval} starting with a $\bigstar$
in this section. The pseudo-code is provided in Appendix~\ref{app:code}, and an overview of our retrieval pipeline is shown in Figure~\ref{fig:query}.

At the core of our approach is the intuition that each local query typically involves relevant entities, like ``\textit{Slytherin}'' and ``\textit{House Cup}'' in the question ``\textit{Has Slytherin won the House Cup?}'', and potential relationships among these entities can guide the retrieval process by identifying the most relevant chunks.
Therefore, we first use the entity extractor, as in the indexing stage, to extract entities from the query, denoted as $\mathcal E_q = \{e_q^1, \cdots e_q^m\}$.
The entities in the query are then mapped to the vertices in our constructed graph. For simplicity, entities that cannot be mapped to any graph vertex are treated as invalid and ignored, as they are likely noise introduced by 
the entity extractor.

{$\bigstar$If no entities are identified, we cannot utilize the entities to support retrieval. In such cases, the query is treated as a global query, and \textit{Dense Retrieval} is performed over the summary tree. We adopt a collapsed-tree dense retrieval approach similar to RAPTOR, leveraging the embedding model used in the indexing stage to encode the query. The similarity between the query embedding and these indexed embeddings is then computed to select the most relevant $k$ chunks as supplements, which are ranked in descending order of similarity.}

Otherwise, since the entity extractor lacks the ability to capture semantic relevance, it often fails to identify the core entities aligned with the query intent, resulting in noisy extractions. 
Simply mapping these entities to the graph is insufficient for filtering out the noise.
Therefore, we introduce a \textit{Graph Filtering} step to retain only the core entities for effective retrieval.
{The key intuition is that query-relevant entities tend to co-occur with each other across the corpus and thus form a locally connected neighborhood in the co-occurrence graph, whereas spurious entities are more likely to be isolated or only weakly connected. Therefore, we use graph proximity as a lightweight proxy for semantic relevance and retain only entity pairs whose shortest-path distance is within $h$ hops.}
Specifically, we enumerate all pairwise combinations of entities from the query as candidate entity pairs.
For each pair, if the two entities are within $h$ hops in the knowledge graph, they are considered semantically related and retained; otherwise, they are discarded as likely irrelevant.
The set of selected entity pairs is denoted as $\mathcal P_h$.
This step is formally defined in Equation~\ref{eq0}, where $\text{Dist}_\mathcal G(\cdot, \cdot)$ returns the hop count of the shortest path between two entities. If no path exists, it returns infinity.
The hyperparameter $h$ controls the strictness of the filtering and can be adaptively adjusted to balance the number of chunks recalled during the following steps.
\begin{equation}
\small
    \mathcal P_h = \left\{(e_q^i, e_q^j) \in \mathcal E_q \times \mathcal E_q \bigm| i<j, \text{Dist}_\mathcal G (e_q^i,e_q^j)\leq h\right\}\label{eq0}
\end{equation}
{$\bigstar$After this filtering step, if no entity pairs meet the criteria, i.e., there are no fine-grained, interrelated entities in the query, which means their relations cannot be extracted within several local chunks. In such cases, we classify it as a coarse-grained global query as well. This also includes cases where the query contains only a single entity, as there are no pair-wise combinations.
However, unlike the previous scenario, entities related to both question and context are still present and can assist in improving chunk selection. To leverage them, we first retrieve the top-$2k$ chunks from the summary tree based on vector similarity as candidate supplementary chunks.
We then apply an \textit{Occurrence Ranking} strategy, ranking these candidate chunks according to the frequency of entity occurrences, defined as $w(c_i) = \text{Count}(c_i, \mathcal E_q)$.
For each candidate summary node, the weight is recursively computed as the sum of the weights of its child nodes, i.e. $w(s_i) = \sum_{c/s\in T_\text{child}(s_i)} w(c/s)$, where $c/s$ may refer to either chunk nodes or summary nodes. This recursive weighting naturally assigns higher scores to high-level summary nodes, aligning with the intuition behind global retrieval.
Finally, we rank the candidate chunks by their computed weights and select the top-$k$ highest-ranked ones as supplementary information.}

\begin{table*}[t]
\caption{Overall results of the three datasets, the \textbf{best results} are in bold and the \underline{runner-up} is underlined. Met. denotes the evaluation metric (accuracy for NovelQA/InfiniteChoice, ROUGE-L for InfiniteQA). IT means indexing time per document, and QT means querying time per question.}
\small
\centering
\label{tab:mainresT}
\resizebox{0.8\textwidth}{!}{
\begin{tabular}{>{\centering\arraybackslash}cccccccc} 
            \toprule
            \multicolumn{2}{c}{Backbone Model} & \multicolumn{3}{c}{Qwen2.5-7B-Instruct} & \multicolumn{3}{c}{Llama3.1-8B-Instruct}\\ 
            \cmidrule{1-8}
            \multicolumn{2}{c}{Dataset}          & NovelQA & InfiniteChoice & InfiniteQA & NovelQA & InfiniteChoice & InfiniteQA\\
            \midrule
\multirow{3}{*}{GraphRAG-L}   
          
& Met. $\uparrow$   & {43.34} & \underline{46.72}  & \underline{13.51}    & \textbf{43.64}      & \textbf{43.66}   &  \underline{6.37}   \\
& IT $\downarrow$   & 13793.89      & 11816.15        & 15686.53          & 4517.09         & 3921.95          & 5533.68 \\
& QT $\downarrow$   & 0.20          & 0.25            & 0.82              & 0.43            & 0.41             & 1.16    \\
            \cmidrule{1-8}                                        
\multirow{3}{*}{{GraphRAG-G}}                                         
& Met. $\uparrow$   & 17.48            & 18.78           & 2.32              & 10.93           & 9.17             &   1.98   \\
& IT $\downarrow$   & 13793.89         & 11816.15        & 15686.53          & 4517.09         & 3921.95         &    5533.68    \\
& QT $\downarrow$   & 15.72            & 16.65           & 2.83              & 3.25            & 3.86             &   3.33   \\                       \cmidrule{1-8}              
\multirow{3}{*}{HippoRAG2}                                        
& Met. $\uparrow$   & \underline{44.60}                & \textbf{48.91}               & 8.43                & 26.03  & 20.69  &    3.58    \\
& IT $\downarrow$   &11102.11                & 10624.65               & 13525.97            & 14279.35              & 17131.67                &    18384.76   \\
& QT $\downarrow$   & 3.81                & 2.40               & 3.10                 & 2.82              & 3.84                &    2.06    \\
            \cmidrule{1-8}   
\multirow{3}{*}{LightRAG}                                            
& Met. $\uparrow$   & 38.57            &{45.41}& 10.41             & 21.82           & 20.52            &  3.44   \\
& IT $\downarrow$   &5290.93           & 4732.98         & 6976.55           & 3416.31         & 3225.94          & 5231.11 \\
& QT $\downarrow$   & 15.68            &  16.03          &  15.97            & 11.44           & 12.92            &  15.44  \\
            \cmidrule{1-8}                                            
\multirow{3}{*}{RAPTOR}                                            
& Met. $\uparrow$   & 37.27            & 34.93           & 6.42              & 40.48           & 37.12            &  5.83   \\
& IT $\downarrow$   &\underline{2847.25}&\underline{2568.26}& \underline{3407.41}    &\underline{2874.65}&\underline{2551.89}&\underline{2844.55} \\
& QT $\downarrow$   & \textbf{0.02}    & \underline{0.08}& \textbf{0.03}     &\textbf{0.02}    & \textbf{0.03}    &\textbf{0.03}  \\
            \cmidrule{1-8}           
\multirow{3}{*}{\ours}           
& Met. $\uparrow$   & \textbf{45.60}   & 43.23          & \textbf{13.65} &\underline{41.26}& \underline{39.74}& \textbf{11.07}\\
& IT $\downarrow$   & \textbf{1397.11} &\textbf{1244.56} & \textbf{1630.87}  & \textbf{1641.49}& \textbf{1433.74} & \textbf{1839.26}\\
& QT $\downarrow$   & \textbf{0.02}    & \textbf{0.05}   & \textbf{0.03}     & \underline{0.03}& \underline{0.05} & \textbf{0.03}\\
            \bottomrule       
            \end{tabular}
}
\end{table*}

If entity pairs exist, i.e., $\mathcal P_h \neq \emptyset$, this indicates the presence of fine-grained relational entities in the query.
In such cases, we perform \textit{Index Mapping}, leveraging the entity-to-chunk index $I_{e\rightarrow c}(\cdot)$ constructed during the indexing stage. Specifically, for each entity pair $(e_q^i,e_q^j)$ in $\mathcal P_h$, we map each entity to the corresponding sets of chunks through the index, and then take their intersection to identify the set of chunks associated with both entities, denoted as $\mathcal C_\text{evidence}^{(e_q^i,e_q^j)}$. $\mathcal C_\text{evidence}$, the union of the $C_\text{evidence}^{(e_q^i,e_q^j)}$ means all the candidate chunks. Formally, we define the \textit{Index Mapping} with Equation~\ref{eq1}.
\begin{align}
\small
    \mathcal C_\text{evidence} &=\bigcup_{(e_q^i, e_q^j) \in \mathcal{P}_h} \mathcal C_\text{evidence}^{(e_q^i,e_q^j)}\\
    &= \bigcup_{(e_q^i, e_q^j) \in \mathcal{P}_h} \left\{I_{e\rightarrow c}(e_q^i) \cap I_{e\rightarrow c}(e_q^j) \right\}\label{eq1}
\end{align}
Once the indexes are mapped, if the number of retrieved chunks does not exceed $k$, we directly return them as the final evidence set.
Otherwise, we first attempt to reduce the number of chunks by decreasing the hop threshold $h$ step-by-step, as tighter structural constraints help eliminate less relevant neighbors.
This continues until the number of retrieved chunks is at most
$k$, including the case where no chunks are retrieved.
If the latter occurs (i.e., the retrieval result becomes empty), we revert to the last non-empty result before the drop and apply an \textit{Entity-Aware Ranking} mechanism to select the top-$k$ chunks from it.
This ranking is based on multiple structural and statistical signals derived during retrieval. Specifically, we compute two metrics for each candidate chunk:
\textbf{Entity Coverage Ranking} counts the number of distinct query-related entities present in the chunk. Chunks covering more entities are prioritized as they are not only more likely to be relevant but also tend to contain more comprehensive contextual information.
\textbf{Entity Occurrence Ranking} ranks the chunks by the total frequency of query-related entities, which is the same as the \textit{Occurrence Ranking}.
Chunks are ranked by these metrics in sequence, first by entity coverage, then by entity occurrence, and the top-$k$ are selected as supplementary evidence. {This operation can be facilitated by the chunk-to-entity index $I_{c\rightarrow e}(\cdot)$ to minimize the cost.}


After retrieving all relevant chunks, we proceed to rank and format the chunks and entities as supplementary input to the LLM.
Following the earlier intuition that entities serve to highlight the key information while chunks provide the supporting details,
we organize the retrieved evidence in an ``entity1-entity2: chunks'' format.
To further reduce token consumption, we apply two optimization strategies.
First, to eliminate redundant input caused by chunks associated with multiple entity pairs, we consolidate these chunks into a single format such as ``entity1-entity2-$\cdots$-entity$n$: chunks''. This ensures that each chunk is included only once, even if it is linked to multiple entity pairs.
Second, we detect and merge consecutive chunks within the evidence set to eliminate overlaps between adjacent chunks.
This chunk merging step further reduces input redundancy and helps minimize token costs.
Finally, we rank the entity pairs based on entity coverage and arrange their corresponding chunks in their original chunk order in the document.


\section{Experiment}

{Due to space limitations, we defer additional results and analysis to the Appendix, including results on different entity extractors (App.~\ref{app:extractors}), results on multihop and domain-specific QA datasets (App.~\ref{app:hotpot}), empirical analysis on the cost of different methods (App.~\ref{app:empirical}), theoretical analysis on the cost of different methods (App.~\ref{app:theory}), and a case study (App.~\ref{app:case}).} We further report fine-grained retrieval ablations
(App.~\ref{app:fine-grained-ablation}),
hop-threshold sensitivity
(App.~\ref{app:hop-sensitivity}),
and the distribution of incorrect answers across retrieval paths
(App.~\ref{app:failure-paths}).

\subsection{Experiment Settings}

We describe our experimental setup, including the choice of base models, datasets, and evaluation metrics. For each component, we detail both the selection criteria and the rationale behind them, aiming to ensure the reproducibility, practicality, and fairness of our evaluation.


\textbf{Base Models.} We adopt two open-source lightweight models, Qwen2.5-7B-Instruct~\citep{qwen} and Llama3.1-8B-Instruct~\citep{llama}, to ensure practicality and reproducibility under limited resources and privacy constraints.
For embeddings, we use BGE-M3~\citep{BGE-M3}, a state-of-the-art open-source model. 
Entity extraction is exemplified by the use of SpaCy, including the ablation study, while comparisons with other extractors are reported in App.~\ref{app:extractors}.


\textbf{Datasets.} We evaluate on QA datasets built from extremely long documents, including NovelQA~\citep{NovelQA} and two subsets of Infinite-Bench~\citep{infiniteBench}, namely InfiniteChoice and InfiniteQA. 
Each document averages about 200k tokens, allowing us to assess global query performance over long contexts (see App.~\ref{Dataset} for details). 
We exclude UltraDomain~\citep{ultradomain} used in LightRAG due to its reliance on LLM-as-judge evaluation~\citep{limitation,calibration,justice}, which raises concerns about reliability. 
Instead, following RAPTOR~\citep{raptor}, we focus on closed-ended QA and multiple-choice tasks for evaluation.


\begin{figure*}[h]
    \centering
    \includegraphics[width=0.85\textwidth]{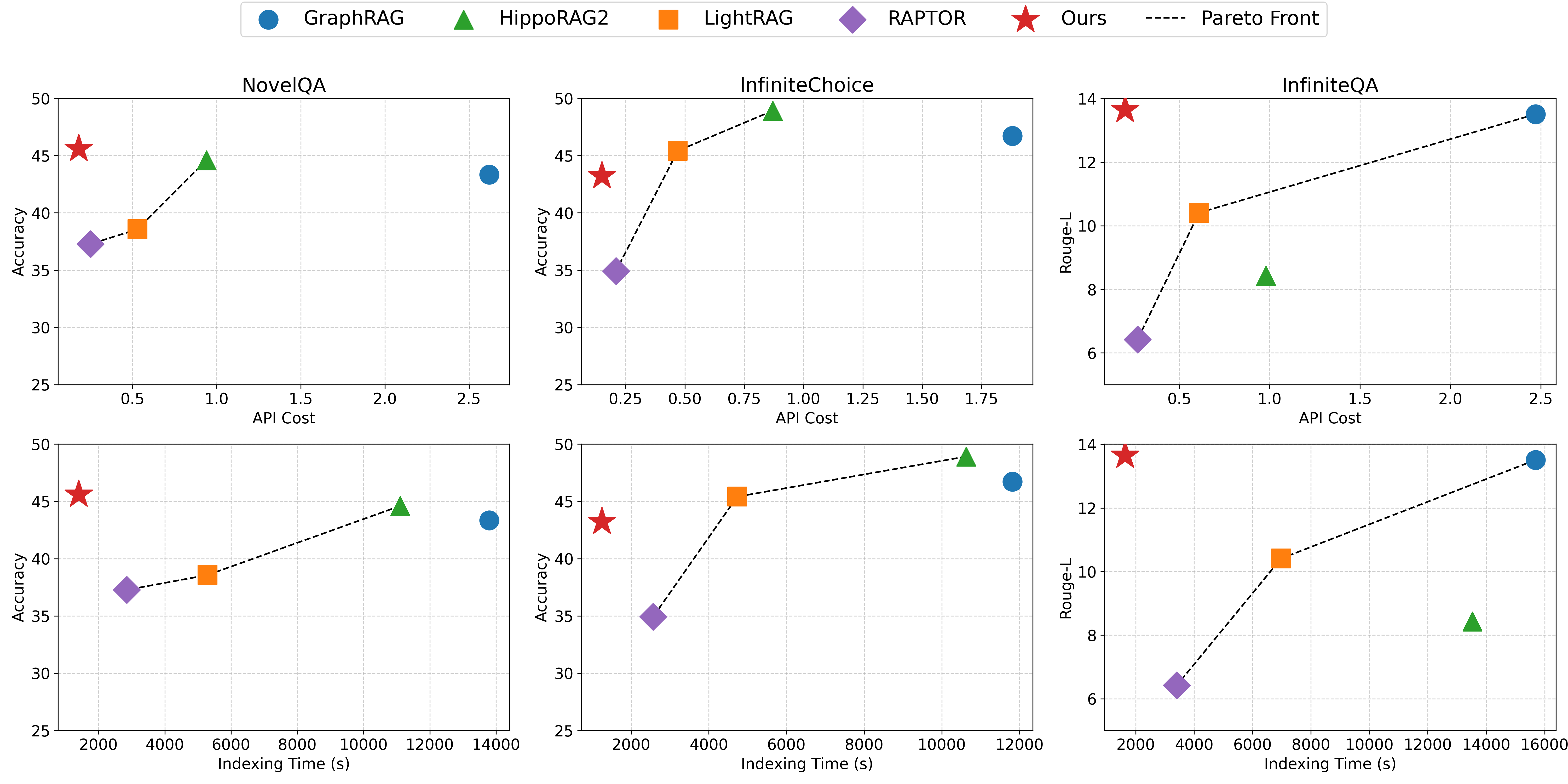}
    \caption{Comparative analysis of our method against baselines based on Qwen across three datasets. Each plot evaluates the trade-off between performance and cost.The black dashed line represents the empirical Pareto front among the evaluated baseline configurations.}
    \label{fig:pareto}
\end{figure*}

\textbf{Metrics.} For multiple-choice and QA tasks, we employ accuracy and ROUGE-L~\citep{rouge-L}, respectively. 
To evaluate efficiency, we measure the average indexing time for each document and the average retrieval latency per query.

\subsection{Baselines}

We compare against all publicly available open-source methods to ensure a comprehensive evaluation, including GraphRAG-Local,
GraphRAG-Global, 
LightRAG-Hybrid, HippoRAG2, and RAPTOR. 
For RAPTOR, we aligned its prompting format with ours, while for LightRAG, HippoRAG2, and GraphRAG, we adopted their default prompts with retries to address JSON extraction failures in smaller models. 
Implementation details are provided in App.~\ref{app:implement}.

\subsection{Experimental Results}

As shown in Table~\ref{tab:mainresT}, \ours\ achieves the highest efficiency in the indexing stage, being \textbf{up to 10$\times$ faster than GraphRAG} and about \textbf{2$\times$ faster than RAPTOR}. In retrieval, \ours\ also shows superior speed, reaching \textbf{over 100$\times$ faster than LightRAG} and nearly \textbf{10$\times$ faster than GraphRAG (local)}. Meanwhile, \ours\ maintains effectiveness on par with GraphRAG, achieving the best performance on NovelQA with Qwen and on InfiniteQA across both backbones. 

In contrast, existing baselines reveal an inherent trade-off between effectiveness and efficiency. GraphRAG achieves the highest QA accuracy, but suffers from extremely low efficiency. LightRAG and HippoRAG2 improve efficiency by simplifying the workflow and reducing some overhead, yet this comes at the cost of decreased QA accuracy. Among the baselines, RAPTOR is the most efficient, but its effectiveness is among the lowest. It's also worth noting that
HippoRAG2 and LightRAG exhibit a substantial decline in performance when the base model switches from Qwen to Llama. We attribute this degradation to the variation in entity extraction capabilities across different LLMs.

In summary, existing methods exhibit trade-offs between efficiency
and effectiveness. As illustrated in Figure~\ref{fig:pareto},
our method (red star) improves the observed Pareto frontier
in most evaluated scenarios, achieving a favorable
cost-performance trade-off. To illustrate how our method works,
we provide a case study in App.~\ref{app:case}.

Furthermore, to better understand the computational burden, we provide an average indexing cost and token usage per book on the three datasets in the App.~\ref {app:empirical}, and estimate the theoretical costs in the App.~\ref{app:theory}.

\subsection{Ablation Study}


\begin{table}[t]
\caption{Ablation study results. The best results for each dataset are highlighted in bold. For other methods, the performance difference compared to \ours\ is shown next to each value, with $\downarrow$ (in red) indicating a decrease and $\uparrow$ (in green) indicating an increase. }
\small
\centering
\label{tab:abstudy}
\resizebox{\linewidth}{!}{
\begin{tabular}{>{\centering\arraybackslash}cccc} 
\toprule
Dataset & {NovelQA} & {InfiniteChoice} & {InfiniteQA} \\
Metric & Acc.  & Acc. & R-L \\
\midrule
\ours &{45.60}   &\textbf{43.23} & \textbf{13.65}  \\
\midrule
Dense Retrieval Only & 42.00{\scriptsize\color{red}~($\downarrow 3.60$)} & 41.04{\scriptsize\color{red}~($\downarrow 2.19$)} &10.03{\scriptsize\color{red}~($\downarrow 3.62$)} \\
\midrule
w/o Graph Filter & 44.30{\scriptsize\color{red}~($\downarrow 1.30$)}  & 36.68{\scriptsize\color{red}~($\downarrow 6.55$)}& 10.47{\scriptsize\color{red}~($\downarrow 3.18$)}\\

w/o Entity-Aware Ranking &44.12{\scriptsize\color{red}~($\downarrow 1.48$)} & 40.17{\scriptsize\color{red}~($\downarrow 3.06$)} & 8.25\ \,{\scriptsize\color{red}~($\downarrow 5.40$)} \\
w/o Both Above & 44.08{\scriptsize\color{red}~($\downarrow 1.52$)} & 35.81{\scriptsize\color{red}~($\downarrow 7.42$)} &10.58{\scriptsize\color{red}~($\downarrow 3.07$)}\\
\midrule
w/o Dense Retrieval  & \textbf{45.90}{\scriptsize\color{green}~($\uparrow 0.30$)}  & 37.99{\scriptsize\color{red}~($\downarrow 5.24$)} & 13.03{\scriptsize\color{red}~($\downarrow 0.62$)}\\
w/o Occurrence Ranking & 44.25{\scriptsize\color{red}~($\downarrow 1.35$)}& 37.99{\scriptsize\color{red}~($\downarrow 5.24$)}& 8.39\ \,{\scriptsize\color{red}~($\downarrow 5.16$)}\\
w/o Both Above & 45.33{\scriptsize\color{red}~($\downarrow 0.27$)} & 37.55{\scriptsize\color{red}~($\downarrow 5.68$)} & 11.07{\scriptsize\color{red}~($\downarrow 2.58$)}\\
\bottomrule
\end{tabular}
}
\end{table}

To thoroughly evaluate the contribution of each component in \ours, we conduct an ablation study on three datasets using the Qwen model. The results are summarized in Table~\ref{tab:abstudy}, which consists of three main sections:

\textbf{Baseline Dense Retrieval Only:} To verify the necessity and effectiveness of our retrieval strategy, we compare \ours\ with a baseline that relies solely on dense retrieval and the built summary tree. The results demonstrate that \ours\ significantly outperforms this baseline, validating the importance of our retrieval enhancements.

\textbf{Local Retrieval Ablations:} To assess the impact of the local retrieval components, we individually and jointly ablate the \textit{Graph Filter} and \textit{Entity-Aware Ranking} modules. Results show that both modules are crucial for local evidence selection. The removal of either leads to a large performance drop, confirming their complementary roles.

\textbf{Global Retrieval Ablations:} Similarly, we evaluate the contribution of the global retrieval by ablating \textit{Dense Retrieval} and \textit{Occurrence Ranking}. Among these, Occurrence Ranking appears more impactful, likely due to its more frequent use in our datasets. Interestingly, we observe an anomalous improvement when removing Dense Retrieval on NovelQA. We hypothesize that this is caused by occasional hallucinations, where the model guesses the correct answer without actual evidence.

\section{Conclusion}

In this paper, we addressed the inefficiency of existing graph-based RAG methods that limits their practicality.
We propose \ours, which streamlines the graph-based RAG pipeline while maintaining competitive QA effectiveness.
For indexing, \ours~recursively built document summary trees with LLMs and efficiently constructed entity-level knowledge graphs using lightweight NLP toolkits such as SpaCy, BERT, substantially reducing time costs and improving practicality. 
For retrieval, we proposed an adaptive strategy that leverages the graph structure to locate relevant chunks and automatically select between local and global retrieval modes, eliminating the need for manually pre-defined query settings.
{Experiments demonstrate that \ours\ achieves up to 10$\times$ faster indexing than GraphRAG while maintaining QA performance.}

\section*{Limitations}

Although we present a streamlined graph-based RAG framework that demonstrates both strong efficiency and effectiveness in this paper, the retrieval design remains relatively intuitive. While we have conducted extensive experiments and explored various alternative retrieval strategies~(some of which are not included in the paper), it is impossible to exhaust all possible retrieval pipeline designs. Therefore, there may still exist more optimal retrieval strategies that could further enhance the performance of our approach.

\section*{Ethical Statement}

Although \ours~aims at retrieving from the knowledge base and providing fact-based answers, this method is still vulnerable to attacks and may generate hallucinations.

\bibliography{custom}
\clearpage
\appendix

\section*{Appendix}
\label{sec:appendix}

\section{Pseudo Code}\label{app:code}
\begin{algorithm}[H]
\caption{The pseudo-code for the retrieval stage.}
\label{alg:retrieval-pseudo}
{\begin{algorithmic}[1]
\REQUIRE $q$, $\mathcal G$, $T$, $k$, $h$
\ENSURE Supplemental text $\mathcal C_s = \{c_1, c_2 \cdots c_n\}, n \leq k$
\STATE entities $\mathcal E_q = \mathtt{SpaCy}(q)$
\IF {Count($\mathcal E_q$) == 0}
\RETURN {$\bigstar$ {$\ \mathcal C_s = \mathtt{DenseRetrieval}(q, k)$}}
\ENDIF
\STATE selected pairs $\mathcal P_h =\mathtt{GraphFilter}(\mathcal E_q,h)$ \  [Eq.~\ref{eq0}]
\IF {Count ($\mathcal P_h$) == 0}
\STATE {$\bigstar\ $candidate chunks $\hat{\mathcal C} = \mathtt{DenseRetrieval}(q,2k)$}
\RETURN {$\bigstar\ $$\mathcal C_s = \mathtt{OccurrenceRank}(\hat{\mathcal C})$}
\ENDIF
\STATE candidate chunks $\hat{\mathcal C} = \mathtt{IndexMapping}(\mathcal P_h)$ \quad [Eq.~\ref{eq1}]
\WHILE {Count($\hat{\mathcal C}$) $> k$}
\STATE $h = h-1$ or $l = l + 1$
\STATE $\hat{\mathcal C}_\text{prev} = \hat{\mathcal C}$
\STATE $\mathcal P_h = \mathtt{GraphFilter}(\mathcal E_q, h)$\  [Eq.~\ref{eq0}]
\STATE $\hat{\mathcal C} = \mathtt{IndexMapping}(\mathcal P_h)$\  [Eq.~\ref{eq1}]
\ENDWHILE
\IF{Count($\hat{\mathcal C}$)==0}
\RETURN $\mathcal C_s = \mathtt{EntityAwareFilter}(\hat{\mathcal C}_\text{prev})$
\ELSE
\RETURN $\mathcal C_s = \hat{\mathcal C}$
\ENDIF
\end{algorithmic}}
\end{algorithm}
\section{Dataset Details}\label{Dataset}

In this section, we provide detailed descriptions of the datasets used in our experiments. While the main paper introduces the overall dataset choices and their relevance to our task, here we include further information on data statistics.

\textbf{NovelQA} has 89 books along with 2305 multiple-choice questions in total, which contain 65 free public-domain books and 24 copyright-protected books purchased from the Internet. It is released with an Apache-2.0 License.
\textbf{InfiniteChoice} has 58 books along with 229 multiple-choice questions in total.
\textbf{InfiniteQA} has 20 books along with 102 questions in total. The InfiniteBench is released with an MIT License.
The links to the two datasets are provided in our code repository, and both datasets are publicly accessible.
We leverage all these datasets for research, aligning with their original purposes.

\begin{figure*}[h]
  \centering
  \subfloat[NovelQA]
  {
      \label{fig:subfig1}\includegraphics[width=0.31\textwidth]{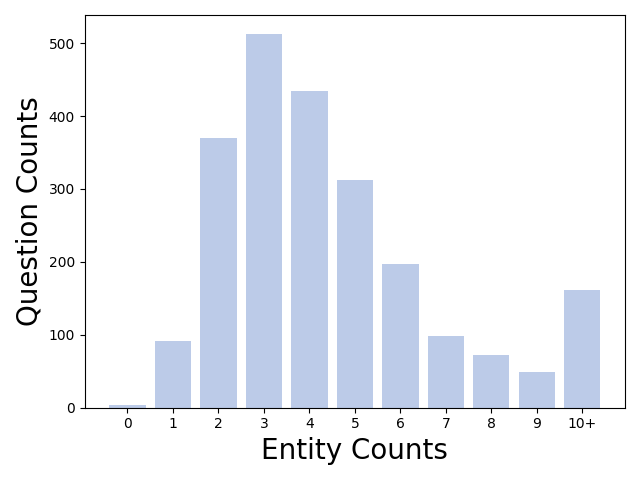}
  }
  \subfloat[InfiniteChoice]
  {
      \label{fig:subfig2}\includegraphics[width=0.31\textwidth]{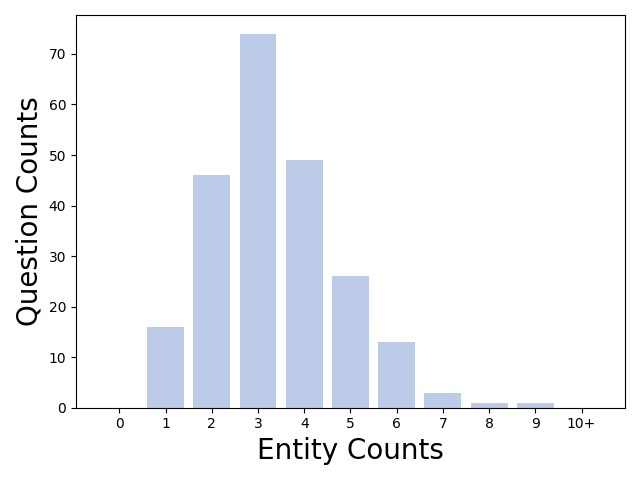}
  }
  \subfloat[InfiniteQA]
  {
      \label{fig:subfig3}\includegraphics[width=0.31\textwidth]{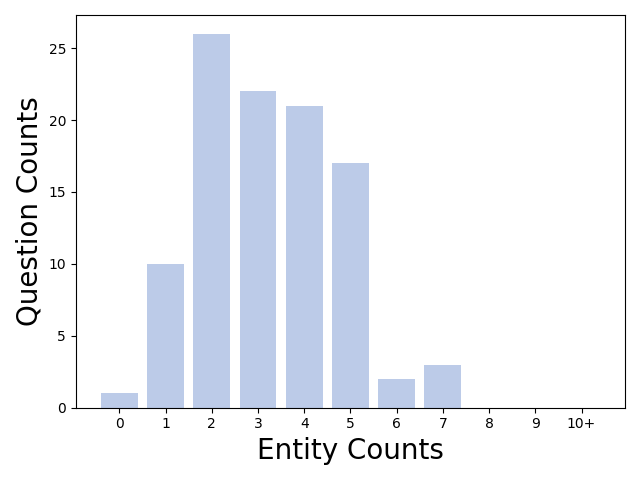}
  }
  \caption{Distribution of questions across different entity count buckets with SpaCy as the extractor.}  
  \label{fig:entity-question}            
\end{figure*}

To provide a deeper insight into our method, we analyze the number of entities involved in each question for all datasets. Specifically, we count the entities mentioned in the question text, excluding those appearing only in the multiple-choice options. The detailed statistics, including the average, minimum, and maximum number of entities per question, are reported in Table~\ref{tab:count}. In addition, Figure~\ref{fig:entity-question} illustrates the distribution of questions across different entity count buckets, offering a clearer view of how entity complexity varies across the datasets.

\begin{table}[t]
\caption{Entity count of each question in each dataset with SpaCy as the extractor.}
\centering
\label{tab:count}
{
\begin{tabular}{>{\centering\arraybackslash}cccc} 
\toprule
Dataset & {NovelQA} & {InfiniteChoice} & {InfiniteQA} \\
\midrule
avg. & 4.60 & 3.24 & 3.37 \\
\midrule
min. & 0 & 1 & 0 \\
\midrule
max. & 24 & 9 & 7 \\
\bottomrule
\end{tabular}
}
\end{table}
\vspace{-0.3cm}





\section{Implementation Details of Baseline Methods}\label{app:implement}

Because of excessive redundant design in the official GraphRAG implementation, we opted for the most widely adopted open-source implementation, nano-GraphRAG~\footnote{\url{https://github.com/gusye1234/nano-graphrag}} for our experiments. 
To adapt GraphRAG for local deployment with HuggingFace models, we utilized the code from LightRAG that supports HuggingFace integration and embedded it into nano-GraphRAG.

For a fair comparison, the hyperparameter settings of all the methods and all baselines are chosen to ensure that the entire pipeline can run on a single NVIDIA A800 GPU with 80GB of memory. 
For the retrieval level of GraphRAG, we choose the best level, i.e., level 2, reported in the corresponding paper.
For the retrieval mode of LightRAG, we choose the Hybrid mode, which is reported as the best mode in the paper.
All LLMs are implemented using the HuggingFace \texttt{transformers} framework, with \texttt{temperature=0.7} and \texttt{max\_new\_tokens=1200} for summarization or entity extraction, except for the GraphRAG, which requires raising the \texttt{max\_new\_tokens} to 8192 for generating a complete JSON structure.
For our method, the configuration files for reproducing the results have been provided on GitHub, with the hyperparameter $k=25$, which is determined by the GPU memory, and $h$ is automatically adjusted during retrieval; therefore, we choose a relatively large value of $4$.





\section{Supplementary Analysis}

\subsection{Comparison between Different Extractors}\label{app:extractors}

\begin{table*}[htbp]
\caption{Overall results with different entity extractors on NovelQA, InfiniteChoice, and InfiniteQA. Met. means the metric for each dataset, and ET means extracting time.}
\centering
\label{tab:extractor}
\resizebox{0.8\textwidth}{!}
{
\begin{tabular}{>{\centering\arraybackslash}cccccccc} 
            \toprule
            \multicolumn{2}{c}{Backbone Model} & \multicolumn{3}{c}{Qwen2.5-7B-Instruct} & \multicolumn{3}{c}{Llama3.1-8B-Instruct}\\ 
            \cmidrule{1-8}
            \multicolumn{2}{l}{Extractor}          & NovelQA & InfiniteChoice & InfiniteQA & NovelQA & InfiniteChoice & InfiniteQA\\
            \midrule                                       
\multirow{2}{*}{NLTK}                                            
& Met. $\uparrow$   & 46.77 & 40.61 & 16.99 & 40.13 & 37.12 & 10.18 \\
& ET $\downarrow$   & 2577.14 & 2315.52 & 3286.00 & 2511.18 & 2241.62 & 3263.51 \\
\cmidrule{1-8}                                            
\multirow{2}{*}{BERT}                                            
& Met. $\uparrow$   & 45.94 & 37.99 & 12.60 & 39.92 & 37.99 &  13.21 \\
& ET $\downarrow$   &58.76 & 60.42 & 84.64 & 63.19 & 60.48 & 84.24 \\
            \cmidrule{1-8}           
\multirow{2}{*}{SpaCy}           
& Met. $\uparrow$   & {45.60}   & 43.23          & {13.65} &{41.26}& {39.74}& {11.07}\\
& ET $\downarrow$   & 39.82 & 35.15 & 45.26 & 38.47 & 33.26 & 46.43 \\
\bottomrule       
            \end{tabular}
}
\end{table*}

To save computing resources and ensure a fair comparison, we do not rebuild the summary tree for different extractors. Instead, we apply each extractor to the same summary tree and report the extraction time and performance in Table~\ref{tab:extractor}. The performance of different extractors is relatively close, and all remain competitive. With the support of modern GPUs, the BERT-based extractor also demonstrates high efficiency. In contrast, the NLTK extractor exhibits low efficiency, since it is primarily a non-industrial toolkit. The SpaCy extractor achieves the highest efficiency while maintaining independence from GPU resources.

\subsection{Results on Multihop and Domain-specific QA Dataset}\label{app:hotpot}

\begin{table}
\centering
\caption{Results on HotpotQA and Medical.}
\resizebox{\linewidth}{!}{\centering
\begin{tabular}{l c c c}
\toprule
\textbf{Method} & \textbf{HotpotQA-Contain} & \textbf{HotpotQA-GPT} & \textbf{Medical-GPT} \\
\midrule
RAPTOR & 55.9 & 58.3 & 55.75 \\
LightRAG & 60.3 & 59.5 & 54.36 \\
HippoRAG & 57.0 & 59.3 & 55.04 \\
\ours & \textbf{61.0} & \textbf{63.9} & \textbf{58.00} \\
\bottomrule
\end{tabular}}
\label{tab:extra}
\vspace{0cm}
\end{table}

\begin{table*}[t]
\caption{Average indexing cost \textbf{per book} and token usage on NovelQA, InfiniteChoice, and InfiniteQA. Costs are computed using the official pricing of the Qwen2.5-7B-Instruct API. For each method, we report absolute values along with relative multiples ($\times$) compared to \ours.}
\centering
\resizebox{0.75\textwidth}{!}{
\centering
\label{tab:cost}
{
\begin{tabular}{>{\centering\arraybackslash}ccccc} 
            \toprule
            \multicolumn{2}{c}{Backbone Model} & \multicolumn{3}{c}{Qwen2.5-7B-Instruct} \\ 
            \cmidrule{1-5}
            \multicolumn{2}{c}{Dataset}          & NovelQA & InfiniteChoice & InfiniteQA \\
            \midrule
\multirow{3}{*}{GraphRAG}   
          
& Input tokens $\downarrow$   & 1,684,445\scriptsize\color{red}~($\times 6.54$)           &1,069,421\scriptsize\color{red}~($\times 4.70$)& 1,396,578\scriptsize\color{red}~($\times 4.80$)       \\
& Output tokens $\downarrow$   &1,778,697\scriptsize\color{red}~($\times 37.55$)           & 1,343,773\scriptsize\color{red}~($\times 32.58$)         & 1,768,985\scriptsize\color{red}~($\times 33.50$)        \\
& Cost $\downarrow$   & 2.62\scriptsize\color{red}~($\times 14.56$)           &  1.88\scriptsize\color{red}~($\times 12.53$)          &  2.47\scriptsize\color{red}~($\times 12.35$)         \\
            \cmidrule{1-5}             
\multirow{3}{*}{HippoRAG2}                                            
& Input tokens $\downarrow$   & 1,236,157\scriptsize\color{red}~($\times 4.80$) & 1,133,410\scriptsize\color{red}~($\times 4.98$) & 1,237,688\scriptsize\color{red}~($\times 4.25$)              \\
& Output tokens $\downarrow$   & 319,988\scriptsize\color{red}~($\times 6.75$) & 297,295\scriptsize\color{red}~($\times 7.21$) & 358,637\scriptsize\color{red}~($\times 6.79$)     \\
& Cost $\downarrow$   & 0.94\scriptsize\color{red}~($\times 5.22$)  & 0.87\scriptsize\color{red}~($\times 5.87$) & 0.98\scriptsize\color{red}~($\times 4.9$)     \\
            \cmidrule{1-5}   
\multirow{3}{*}{LightRAG}                                            
& Input tokens $\downarrow$   & 713,782\scriptsize\color{red}~($\times 2.77$)            &621,740\scriptsize\color{red}~($\times 2.73$)& 795,834\scriptsize\color{red}~($\times 2.74$)              \\
& Output tokens $\downarrow$   &173,186\scriptsize\color{red}~($\times 3.66$)          & 156,146\scriptsize\color{red}~($\times 3.78$)        & 212,674\scriptsize\color{red}~($\times 4.03$)       \\
& Cost $\downarrow$   & 0.53\scriptsize\color{red}~($\times 2.94$)            &  0.47\scriptsize\color{red}~($\times 3.13$)          &  0.61\scriptsize\color{red}~($\times 3.05$)          \\
            \cmidrule{1-5}    
\multirow{3}{*}{RAPTOR}                                            
& Input tokens $\downarrow$   & 323,233\scriptsize\color{red}~($\times 1.26$)            &277,796\scriptsize\color{red}~($\times 1.22$)& 362,129\scriptsize\color{red}~($\times 1.24$)              \\
& Output tokens $\downarrow$   &88,889\scriptsize\color{red}~($\times 1.88$)           & 71,582\scriptsize\color{red}~($\times 1.74$)         & 92,138\scriptsize\color{red}~($\times 1.74$)        \\
& Cost $\downarrow$   & 0.25\scriptsize\color{red}~($\times 1.39$)            &  0.21\scriptsize\color{red}~($\times 1.40$)          &  0.27\scriptsize\color{red}~($\times 1.35$)          \\
            \cmidrule{1-5}           
\multirow{3}{*}{\ours}           
& Input tokens $\downarrow$   & 257,500            &227,474& 290,923              \\
& Output tokens $\downarrow$   &47,375           & 41,244         & 52,800        \\
& Cost $\downarrow$   & 0.18            &  0.15          &  0.20          \\
            \bottomrule       
            \end{tabular}
}}
\end{table*}

We conduct additional experiments on the HotpotQA dataset (randomly sampling 1,000 entries from the validation set) and a domain-specific dataset from GraphRAG-Bench~\citep{medical}, namely Medical. All baseline methods share the same backbone model, Llama3-8B-Instruct, to ensure fair comparison. For HotpotQA, we report Contain-Match (HotpotQA-Contain), and GPT-Evaluation Accuracy (HotpotQA-GPT), while for Medical, we use GPT-Evaluation Accuracy (Medical-GPT) only, as this dataset poses greater challenges for automatic evaluation.

As shown in Table~\ref{tab:extra}, even on HotpotQA, where documents do not have an inherent sequential order, our method still clearly outperforms the baselines, demonstrating its generality. Moreover, on the domain-specific Medical dataset, our method also achieves strong performance. Taken together, the results on these two datasets highlight the strong generalization ability of our approach.

\subsection{Empirical Comparison on Method Cost}\label{app:empirical}

To assess the computational overhead of our method compared to baseline approaches, we conduct theoretical analyses in the following section. 

In addition to the wall-time comparison reported in Table~\ref{tab:mainresT}, Table~\ref{tab:cost} summarizes the average indexing cost and token usage per book on NovelQA, InfiniteChoice, and InfiniteQA, calculated using the official pricing of the Qwen2.5-7B-Instruct API. 
For clarity, we report both the absolute values and relative multiples compared to \ours, which provides further evidence of the superior efficiency of our approach. 
Overall, \ours\ substantially reduces indexing expense, e.g., over $12\times$ lower cost than GraphRAG and around $3\times$ lower than LightRAG, while achieving comparable or better QA performance.

\begin{figure*}[t]
  \centering
  \subfloat[NovelQA]
  {
      \label{fig:NovelQA}\includegraphics[width=0.48\textwidth]{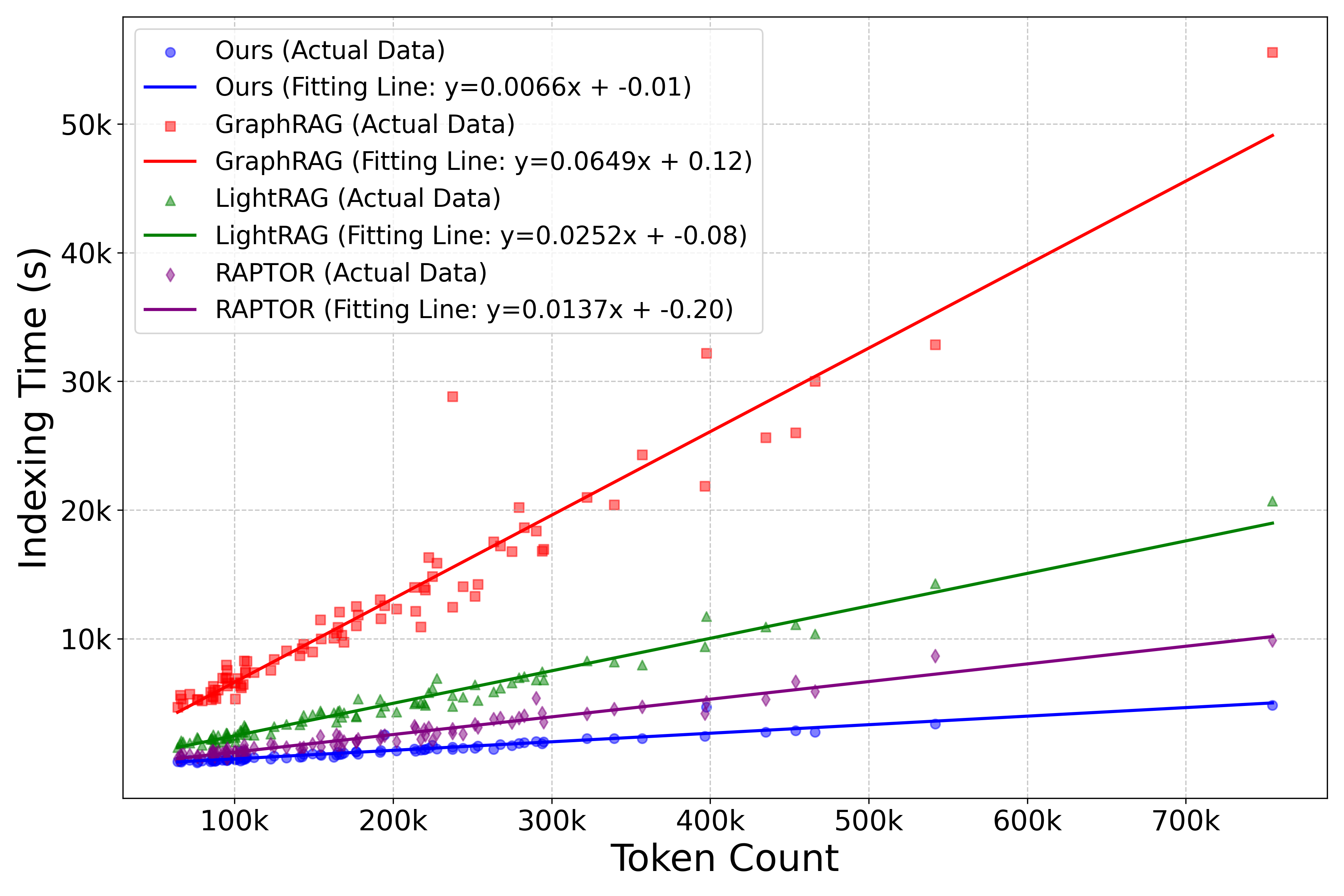}
  }
  \subfloat[InfiniteChoice]
  {
      \label{fig:InfiniteChoice}\includegraphics[width=0.48\textwidth]{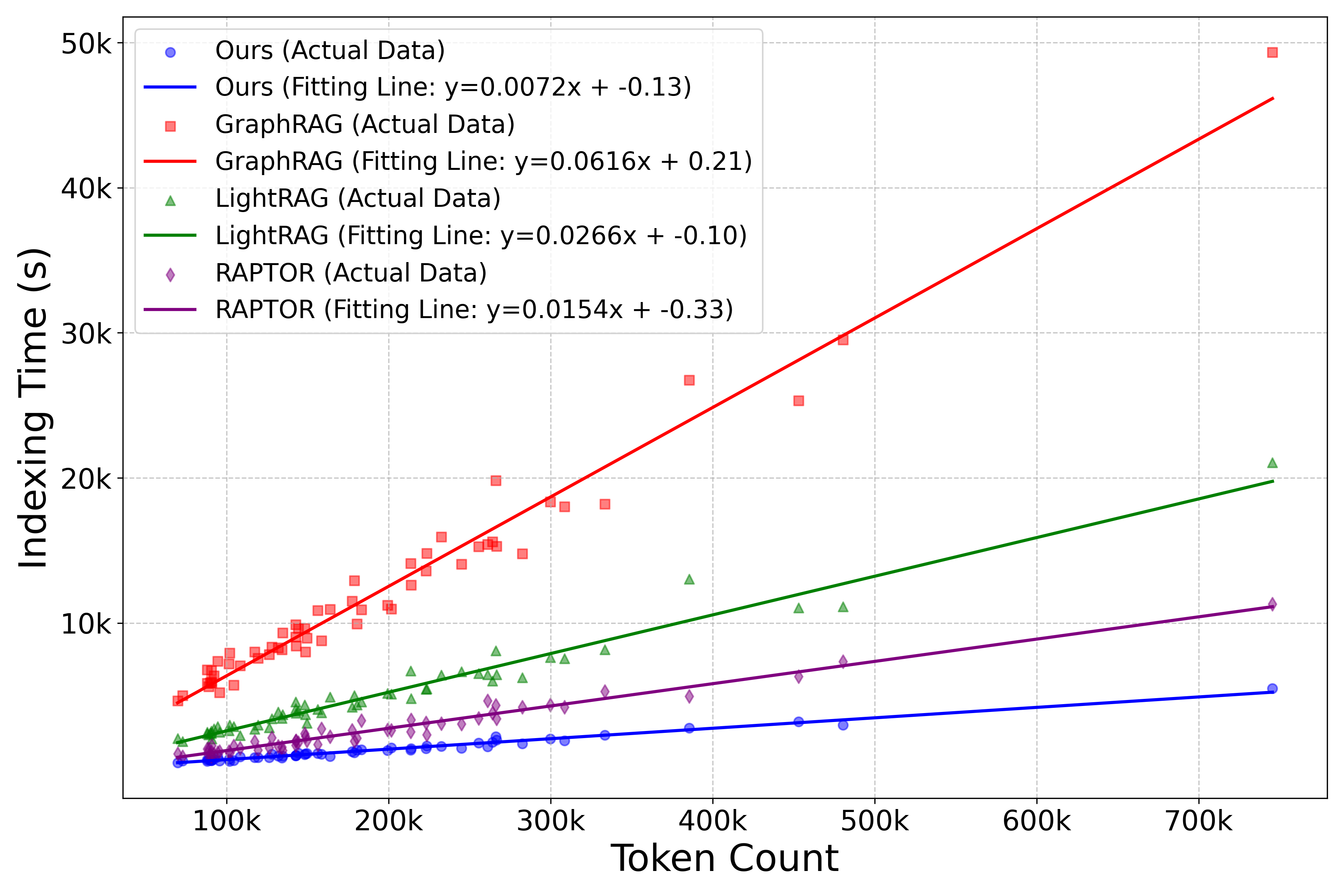}
  }
  \caption{Time cost as a function of document token count for each method. The statistic is based on NovelQA and InfiniteChoice with Qwen as the base model.}\label{fig:token-time}   
\end{figure*}

Furthermore, to provide a more intuitive comparison of how indexing time scales with the size of the text corpus, we fitted a function to the indexing time versus text length data for all four methods, as shown in Figure~\ref{fig:token-time}. The $R^2$ values for each method on every dataset exceed 0.90, indicating a strong goodness-of-fit. From the figure, it is evident that all four methods exhibit approximately linear growth, while our method demonstrates \textbf{the lowest slope}. This observation aligns with both our theoretical analysis and experimental results, providing clear evidence that our approach \textbf{scales linearly with the knowledge base size at the minimal rate among the compared methods.}

\subsection{Theoretical Comparison on Method Cost}\label{app:theory}

\subsubsection{Theoretical Indexing Cost}\label{app:index}

Following the symbols defined in Table~\ref{tab:cost_theoretical}, we use $n$ to denote the number of chunks, $m$ for the number of communities detected by GraphRAG, and $g$ to represent the group size (or maximum group size) for \ours\ and RAPTOR. Furthermore, since the dominant overhead of each method comes from LLM invocations, we next analyze, method by method, the number of LLM calls required during the indexing stage.

\textbf{\ours:} Our method builds a tree with $n$ leaf nodes and each non-leaf node has $g$ child nodes. The number of LLM calls is equal to the number of non-leaf nodes. The non-leaf nodes can be listed by level and form a geometric sequence with the first term $\lceil \frac{n}{g} \rceil$ and the common ratio $\frac{1}{g}$, where each term is rounded up to the nearest integer. 
Consequently, the total number of LLM calls is:
$$S = \sum_{k=1}^K\Big\lceil\frac{n}{g^k}\Big\rceil,$$
where $K = \lceil \log_g n \rceil$.
Since $\lceil x \rceil \in [x, x+1)$, the upper and lower bounds of the equation above can be derived as:
$$\sum_{k=1}^K \frac{n}{g^k} \leq S < \sum_{k=1}^K \frac{n}{g^k} + K$$
Applying the geometric series sum formula $S = a\frac{1-r^n}{1-r}$, the equation above can be rewritten as:
$$\frac{n}{g-1}\left(1- g^{-K}\right) \leq S < \frac{n}{g-1}\left(1- g^{-K}\right) + K$$
Because $K$ grows only logarithmically with respect to $n$, its contribution to the overall number of LLM calls is negligible. For simplicity, we therefore approximate the count by ${n}/{(g-1)}$.

\textbf{RAPTOR:} Similar to our method, RAPTOR builds a tree with $n$ leaf nodes. However, each non-leaf node has at most $g$ child nodes, resulting in the number of non-leaf nodes being larger than \ours. Therefore, the \textbf{lower bound} is ${n}/({g-1})$. Compared with RAPTOR, it is worth noting that our efficiency gain further comes from avoiding clustering: each group in our method contains exactly $g$ chunks, whereas RAPTOR forms groups of at most $g$ chunks. As a result, our summary tree is flatter with fewer nodes, while RAPTOR tends to produce a deeper tree with more intermediate nodes, leading to higher indexing cost.

\textbf{LightRAG:} LightRAG extracts the entities and relations from each chunk and then assembles them into an entity graph. The extraction phase takes $n$ LLM calls, and the assembling phase does not need LLM calls. Therefore, it requires $n$ times as many LLM calls in total.

\textbf{GraphRAG:} GraphRAG extracts the entities and relations from each chunk and formats an entity graph. Then, GraphRAG clusters the nodes and summarizes each community to aggregate the information using an LLM. Therefore, the extraction phase evokes $n$ LLM calls, and the summarization phase calls the LLM $m$ times, which is $n+m$ in total.

\textbf{HippoRAG2:} HippoRAG2 extracts the entities first and then extracts relations from each chunk. It takes $2n$ LLM calls. It's worth noting that HippoRAG2 does not generate the tedious entity descriptions or relation descriptions, which significantly reduces the token cost compared to LightRAG and GraphRAG.

\subsubsection{Theoretical Querying Cost}\label{app:query}

\begin{table}
\centering
\caption{Comparison of theoretical LLM calls across methods, where $n$ is the number of chunks and $m$ is the community counts for GraphRAG. $C_\text{window}$ is the length of LLMs' context window.}
\resizebox{\linewidth}{!}{\centering
\begin{tabular}{l c c}
\toprule
\textbf{Method} & \textbf{Indexing} & \textbf{Retrieval} \\
\midrule
GraphRAG & $n+m$ & $m\times \text{len}(m)/C_\text{window}$ \\
LightRAG & $n$ & $1$ \\
HippoRAG2 & $2n$ & $1$ \\
RAPTOR   & $\geq \lceil{n}/({g-1})\rceil$ & $0$ \\
\textbf{\ours} & $\lceil{n}/({g-1})\rceil$ & $0$ \\
\bottomrule
\end{tabular}}
\label{tab:cost_theoretical}
\vspace{0cm}
\end{table}

Following the symbols defined in Section~\ref{sec:method}, we use $n$ to denote the number of chunks, $m$ for the number of communities detected by GraphRAG, $g$ for the maximum group size, and $k$ to represent the maximum number of chunks to retrieve. The overall comparison is provided in Table~\ref{tab:cost_theoretical}.

\textbf{\ours} In the query stage, the primary computational overhead arises from entity extraction, whose cost varies with the efficiency of the extractor employed. In comparison, the graph search introduces only minor latency, which remains negligible relative to the time consumed by LLM invocations.

\textbf{RAPTOR:} The primary computational overhead of RAPTOR lies in the dense embedding retrieval, which relies on GPUs for acceleration.

\textbf{LightRAG:} LightRAG uses the LLM to extract entities from the query; hence, the overhead is dominated by the LLM invocation.

\textbf{GraphRAG:} In local mode, GraphRAG performs dense retrieval to identify the most similar nodes. In global mode, it relies on the LLM to select relevant communities for answering the query, resulting in $m\times \text{len}(m) / C_\text{window}$ calls.

\textbf{HippoRAG2:} HippoRAG2 employs the LLM for reranking, so the query-time cost is driven by LLM calls.

\subsection{Case Study}\label{app:case}

\begin{figure*}[t]
  \centering
  \subfloat[Local Search]
  {
      \label{fig:localcase}\includegraphics[width=0.48\textwidth]{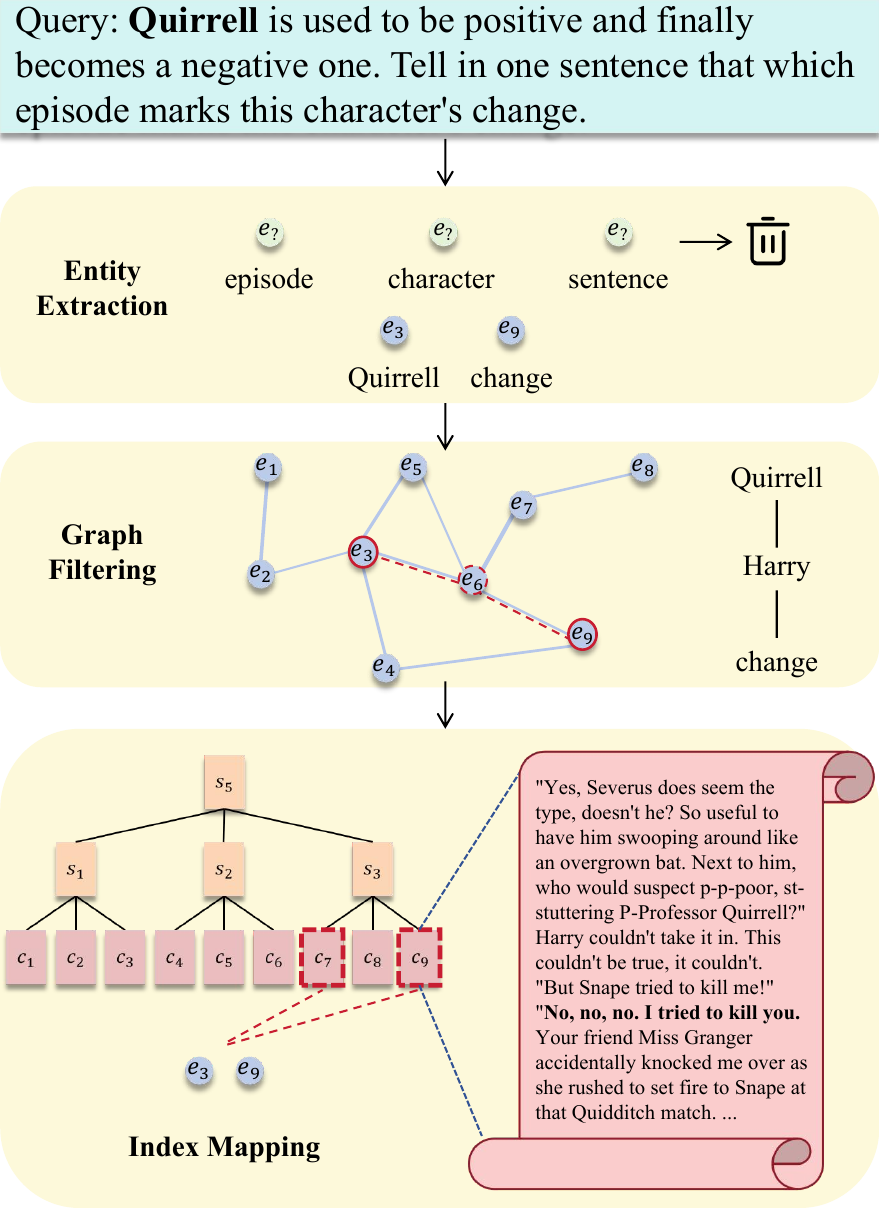}
  }
  \subfloat[Global Search]
  {
      \label{fig:globalcase}\includegraphics[width=0.48\textwidth]{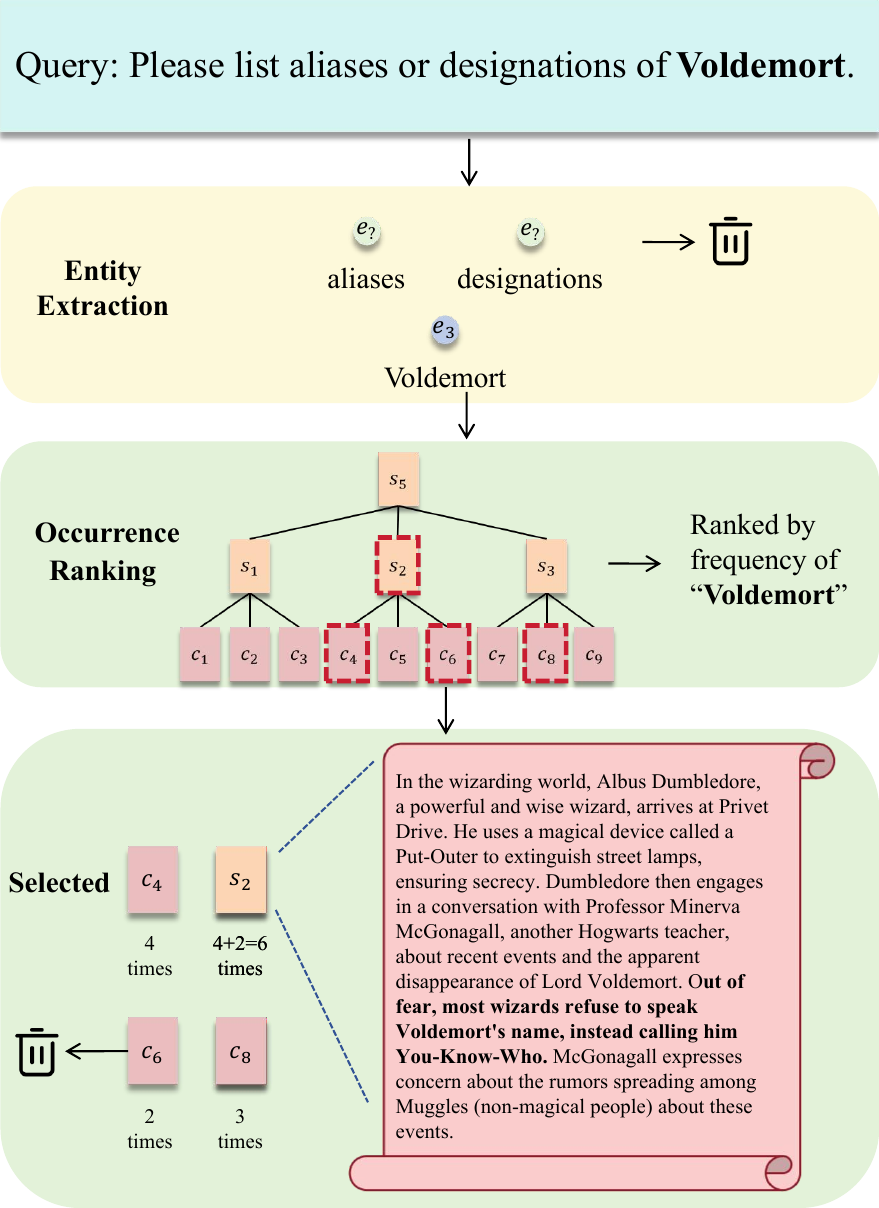}
  }
  \caption{Case study examples: local search (left) vs. global search (right).}\label{fig:casestudy}   
\end{figure*}

To better illustrate how our method operates, we present two case studies under different query modes, as shown in Figure~\ref{fig:casestudy}. In both cases, named entities are first extracted and then mapped onto the graph constructed during the indexing stage. For the case on the left, among the five extracted entities, three are absent from the graph, while two are connected. We identify the shortest path, \textit{Quirrell - Harry - Change}, which enables us to localize the relevant chunks based on these entities. This leads us to chunks simultaneously associated with \textit{Quirrell} and Change, from which we uncover key evidence that \textit{Quirrell} attempted to kill Harry during the Quidditch match.

For the case on the right, only one of the three entities appears in the graph. Hence, we employ the global query mode, performing dense retrieval over the entire summary tree and ranking candidates by the frequency of the entity Voldemort. 
Voldemort occurs four times in $c_4$, twice in $c_6$,
and three times in $c_8$, while the occurrence weight of
$s_2$ is the sum over its children:
$w(s_2)=w(c_4)+w(c_5)+w(c_6)=4+0+2=6$.
We ultimately select $c_4$ and $s_2$, and within the summary of $s_2$, we identify an alias of Voldemort generated by the LLM.

\subsection{Fine-Grained Retrieval Ablations}
\label{app:fine-grained-ablation}

We further evaluate individual retrieval components using
Qwen2.5-7B-Instruct on all 2,305 questions from NovelQA.
Following a CUDA environment update, we reran the full pipeline
and all ablated variants under the same updated environment.
We report this experiment separately from the original results,
using the rerun full pipeline as the reference for all comparisons
in Table~\ref{tab:fine-grained-ablation}.

\begin{table}[t]
\centering
\small
\caption{Fine-grained ablations on NovelQA with
Qwen2.5-7B-Instruct. All variants were evaluated under the
same updated environment. QT denotes average query time.}
\label{tab:fine-grained-ablation}
\resizebox{\columnwidth}{!}{%
\begin{tabular}{lrr}
\toprule
Variant & Acc.\ (\%) $\uparrow$ & QT (s) $\downarrow$ \\
\midrule
Full Pipeline                & \textbf{46.16} & 0.0181 \\
w/o Entity Coverage Ranking  & 45.60 & 0.0179 \\
w/o Entity Occurrence Ranking & 45.42 & 0.0202 \\
w/o Occurrence Ranking       & 44.77 & 0.0190 \\
w/o Hop Adjustment           & 45.29 & \textbf{0.0176} \\
w/o Chunk Merging            & 38.22 & 0.0441 \\
w/o Entity Formatting        & 44.77 & 0.0183 \\
\bottomrule
\end{tabular}%
}
\end{table}

Among the individually ablated components, chunk merging has
the largest observed impact: removing it reduces accuracy by
7.94 percentage points and increases average query time from
0.0181 to 0.0441 seconds.
Removing the other components leads to smaller accuracy drops,
ranging from 0.56 to 1.39 percentage points.
These results identify chunk merging as particularly important
in this setting, while the ranking, hop-adjustment, and formatting
components provide additional accuracy gains.

\subsection{Sensitivity to the Hop Threshold}
\label{app:hop-sensitivity}

We examine the effect of the hop threshold using
Qwen2.5-7B-Instruct on all 2,305 NovelQA questions.
For each $h \in \{2,3,4,5,6\}$, we compare adaptive hop
adjustment, initialized at $h$, against a variant that keeps
the threshold fixed at the same value.

\begin{table}[t]
\centering
\small
\caption{NovelQA accuracy (\%) with and without hop adjustment
across five initial hop thresholds.}
\label{tab:hop-sensitivity}
\resizebox{\columnwidth}{!}{%
\begin{tabular}{lrrrrr}
\toprule
Initial $h$ & 2 & 3 & 4 & 5 & 6 \\
\midrule
w/ Hop Adjustment
& \textbf{45.55} & \textbf{45.68} & \textbf{46.16}
& \textbf{45.90} & \textbf{45.99} \\
w/o Hop Adjustment
& 45.47 & 44.82 & 45.29 & 45.08 & 45.55 \\
\bottomrule
\end{tabular}%
}
\end{table}

As shown in Table~\ref{tab:hop-sensitivity}, adaptive hop
adjustment improves accuracy at every evaluated threshold,
with gains ranging from 0.08 to 0.87 percentage points.
Across these settings, the accuracy range is 0.61 percentage
points with adjustment, compared with 0.73 without adjustment.
This indicates slightly lower sensitivity to the initial
threshold in the evaluated setting.

\subsection{Failure Distribution across Retrieval Paths}
\label{app:failure-paths}

We analyze the retrieval paths taken by the 1,241 incorrectly
answered NovelQA questions in the full-pipeline run with
Qwen2.5-7B-Instruct.
Table~\ref{tab:failure-paths} groups these errors by retrieval
path and the recorded number of hop-adjustment iterations.
The percentages describe the composition of the error set,
rather than the error rate within each retrieval path.

\begin{table}[t]
\centering
\small
\caption{Distribution of the 1,241 incorrect answers on NovelQA
across retrieval paths. Iterations denote hop-adjustment
iterations; shares are percentages of all incorrect answers.}
\label{tab:failure-paths}
\resizebox{\columnwidth}{!}{%
\begin{tabular}{lrrr}
\toprule
Retrieval Path & Iterations & Errors & Share (\%) \\
\midrule
Entity-Aware Ranking & 3 & 698 & 56.2 \\
Local Retrieval      & 0 & 358 & 28.8 \\
Occurrence Ranking   & -- & 118 & 9.5 \\
Local Retrieval      & 2 & 55 & 4.4 \\
Entity-Aware Ranking & 2 & 11 & 0.9 \\
Dense Retrieval      & -- & 1 & 0.1 \\
\midrule
Total                & -- & 1,241 & 100.0 \\
\bottomrule
\end{tabular}%
}
\end{table}

The largest group comprises 698 errors on the entity-aware
ranking path after three hop-adjustment iterations, accounting
for 56.2\% of all incorrect answers.
A further 358 errors occur on the local retrieval path without
hop adjustment, while 118 occur on the occurrence-ranking path.
These concentrations motivate closer inspection of evidence
retention during filtering, entity-based chunk localization,
and entity matching in fallback retrieval.
Path membership alone does not distinguish retrieval failures
from answer-generation errors.

\section{The Use of LLMs}

This paper employed LLMs solely for grammatical correction and stylistic refinement, with the purpose of more effectively communicating our results and conclusions.

\end{document}